\documentclass[journal,twoside,web]{ieeecolor}
\usepackage{generic}
\usepackage{cite}
\usepackage{amsmath,amssymb,amsfonts}
\usepackage{algorithmic}
\usepackage{graphicx}
\usepackage{algorithm,algorithmic}
\usepackage{hyperref}
\hypersetup{
  colorlinks=true,
  linkcolor=black,
  citecolor=black,
  urlcolor=black
}
\usepackage{textcomp}
\usepackage{colortbl}

\usepackage{orcidlink}

\usepackage{adjustbox}
\usepackage[font=footnotesize]{caption}
\usepackage{amsmath,amsfonts}
\usepackage{algorithmic}
\usepackage{algorithm}
\usepackage{array}
\usepackage[caption=false,font=normalsize,labelfont=sf,textfont=sf]{subfig}
\usepackage{textcomp}
\usepackage{stfloats}
\usepackage{url}
\usepackage{verbatim}
\usepackage{graphicx}
\usepackage{cite}

\usepackage{float} 
\usepackage{placeins}

\usepackage{tabularx}
\usepackage{booktabs}
\usepackage{graphicx}
\usepackage{array}
\usepackage{threeparttable}

\usepackage{caption}
\usepackage{multirow}
\usepackage{booktabs}
\usepackage{array}
\usepackage{xcolor}

\def\BibTeX{{\rm B\kern-.05em{\sc i\kern-.025em b}\kern-.08em
    T\kern-.1667em\lower.7ex\hbox{E}\kern-.125emX}}
\begin{document}
\title{SOC-DGL: Social Interaction Behavior Inspired Dual Graph Learning Framework for Drug-Target Interaction Identification}


\author{Xiang Zhao \orcidlink{0009-0002-4481-7429}, Ruijie Li \orcidlink{0009-0009-1208-4399}, \IEEEmembership{Student Member, IEEE}, Qiao Ning
\orcidlink{0000-0002-4809-5104}, Shikai Guo \orcidlink{0000-0002-8554-6365}, Hui Li \orcidlink{0000-0003-1923-0669} and Qian Ma
\thanks{This work was supported by the National Natural Science Foundation of China (62302075). 
\textit{(Xiang Zhao and Ruijie Li contributed equally and are co-first authors. Corresponding authors: Qiao Ning; Shikai Guo.)}}
\thanks{Xiang Zhao is with the College of Marine Electrical Engineering, Dalian Maritime University, Dalian 116026, China (e-mail: zhaoxiang@dlmu.edu.cn). }
\thanks{Ruijie Li is with the AI thrust, Information Hub, Hong Kong University of Science and Technology (GuangZhou), Guangzhou 511453, China (e-mail: lrj\_@dlmu.edu.cn).}
\thanks{Qiao Ning is with the School of Artificial Intelligence and Computer Science, Jiangnan University, Wuxi 214122, China (e-mail: ningq669@jiangnan.edu.cn).}
\thanks{Shikai Guo, Hui Li and Qian Ma are with the Department of Information Science and Technology, Dalian Maritime University, Dalian 116026, China (e-mail: Shikai.guo@dlmu.edu.cn, li\_hui@dlmu.edu.cn, maqian@dlmu.edu.cn).}
\thanks{The source code and data for SOC-DGL are available for academic purposes at \url{https://github.com/Zhaoxiang0422/SOC-DGL}.}}

\maketitle

\begin{abstract}
The identification of drug-target interactions (DTI) is critical for drug discovery and repositioning, as it reveals potential therapeutic uses of existing drugs, accelerating development and reducing costs. \textcolor{red}{However, most existing models focus only on direct similarity in homogeneous graphs, failing to exploit the rich similarity in heterogeneous graphs.} \textcolor{red}{To address this gap, inspired by real-world social interaction behaviors, we propose SOC-DGL, which comprises two specialized modules: the Affinity-Driven Graph Learning (ADGL) module, learning global similarity through an affinity-enhanced drug-target graph, and the Equilibrium-Driven Graph Learning (EDGL) module, capturing higher-order similarity by amplifying the influence of even-hop neighbors using an even-polynomial graph filter based on balance theory. This dual approach enables SOC-DGL to effectively capture similarity information across multiple interaction scales within affinity and association matrices.} To address the issue of imbalance in DTI datasets, we propose an adjustable imbalance loss function that adjusts the weight of negative samples by the parameter. Extensive experiments on four benchmark datasets demonstrate that SOC-DGL consistently outperforms existing state-of-the-art methods across both balanced and imbalanced scenarios. \textcolor{red}{Moreover, SOC-DGL successfully predicts the top 9 drugs known to bind ABL1, and further analyzed the 10th drug, which has not been experimentally confirmed to interact with ABL1, providing supporting evidence for its potential binding.}
\end{abstract}


\section{Introduction}
\label{sec:introduction}
\IEEEPARstart{D}{rug} 
repurposing, the strategy of applying approved medications to novel therapeutic indications, has gained prominence due to the stagnation in traditional drug discovery~\cite{jourdan2020drug}. 
Drug-target interactions (DTI) are central to pharmacology, yet conventional drug discovery remains time-consuming, costly, and high-risk compared to repurposing~\cite{farha2019drug}. Consequently, computational methods have emerged to predict novel DTI strengths. With the exploding of biomedical data, artificial intelligence-based methods for drug repurposing have proliferated~\cite{ an2025multi}, including two key categories: machine learning-based methods and network-based methods.


 Predicting DTI typically involves mapping drug-target pairs to feature vectors and applying machine learning for classification~\cite{zhao2024msi}. 
However, these machine learning-based methods are limited by the need for positive and negative samples and are susceptible to sample imbalance, which can introduce noise and reduce prediction accuracy. \textcolor{red}{Neural networks have been widely applied in various fields~\cite{movassagh2023artificial, yadawad2024auto, alzubi2025multimodal, lei2022evennet, le2023predicting, zhao2022improved, li2025mhmda}.}
Network-based methods typically build homogeneous or heterogeneous biological networks, infer interactions from drug-drug similarity, target-target similarity, or DTI~\cite{hua2025mmdg}. 
Although network-based methods are effective to some extent, they primarily focus on direct interactions and fail to explore similarity information embedded in heterogeneous information networks.

\textcolor{red}{Most existing DTI prediction methods primarily focus on mining direct similarity information within homogeneous graphs, neglecting the potentially rich high-order similarity information embedded in heterogeneous drug-target networks~\cite{shi2024review}. This oversight limits the ability to fully capture complex and indirect relationships essential for accurate DTI identification, especially under data imbalance scenarios. Therefore, there is a clear research gap in designing effective models that can comprehensively exploit both direct and high-order similarity information from heterogeneous graphs.}

To address this challenge, this study proposes a novel social interaction-inspired dual-enhanced graph reasoning mechanism, termed SOC-DGL, whose main contributions are:

\begin{itemize}
\item Inspired by real-world social interactions, this study proposes two complementary modules, ADGL and EDGL, enabling a comprehensive exploration of homogeneous information for both drugs and targets across multiple interaction scales. Based on ADGL, which is built upon an extensive social interaction framework to achieve global representation learning on affinity-enhanced drug-target networks, EDGL further incorporates balance theory and employs an even-polynomial graph filter to capture high-order even-hop neighbor relationships in the heterogeneous drug-target graph.

\item To mitigate the impact of imbalanced data on SOC-DGL, this study designs an adjustable imbalance loss function that weights negative samples and introduces an adjustable parameter \( \varpi \). Extensive experimental results on imbalanced benchmark datasets consistently demonstrate the effectiveness.

\item ABL1, a key pathogenic target in Chronic Myeloid Leukemia, is effectively targeted by 10 drugs predicted by SOC-DGL. Utilizing a homogeneous information aggregation mechanism, SOC-DGL uncovers a potential association between Debromohymenialdisine and ABL1, highlighting its capability to identify novel DTI.
\end{itemize}

\textcolor{red}{
The objective of this study is to explore rich similarity information from heterogeneous drug-target graphs by a dual graph learning framework inspired by social interaction strategies, thereby enhancing the accuracy of DTI prediction. The rest of this article is organized as follows. Section II reviews related work to SOC-DGL. Section III presents the datasets and details the SOC-DGL framework. Section IV describes the experimental settings and discusses results. Finally, Section V concludes the paper and outlines future research directions.}

\section{Related Work}
\textcolor{red}{SOC-DGL is closely tied to previous studies for DTI prediction, particularly those related to DTI prediction based on similarity and high-order similarity information learning.}

\subsection{DTI Prediction Based on Similarity}
The integration of similarity information plays a pivotal role in advancing DTI prediction models. Traditionally, similarity mining has focused on isomorphic networks, where pairwise similarity between drugs and targets is utilized. \textcolor{red}{DTiGEMS+~\cite{thafar2020dtigems+}, combines graph embedding, mining, and similarity techniques for improved drug–target interaction prediction. iNGNN-DTI~\cite{sun2024ingnn}, an interpretable graph neural network, integrates pre-trained models and enhances accuracy while offering insights into DTI.} However, recent work has shifted towards heterogeneous graphs, which integrate multiple types of relationships. \textcolor{red}{Shao et al. introduced DTI-HETA~\cite{shao2022dti}, an end-to-end model that integrates heterogeneous graphs and attention mechanisms for improved DTI prediction. Notably, GIAE-DTI~\cite{10676326} combines cross-modal information from drugs and targets with self-supervised learning, demonstrating promising performance.}  However, these methods primarily rely on information derived from traditional similarity matrices, overlooking the rich and potentially valuable indirect similarity information between drugs and targets embedded within heterogeneous graphs. Such higher-order similarity information is critical for predicting potential DTI associations.


\subsection{High-order Similarity Information Learning}
Effectively learning high-order similarity information aids in uncovering complex higher-order isomorphic relationships in graph data. In HSC~\cite{peng2022multi}, high-order similarity is learned from relationships between dimensional spaces and jointly optimized with ordinary similarity for clustering. HSL~\cite{mi2024unsupervised} learns high-order similarity by considering both first-order and high-order similarities in a low-dimensional embedding space. \textcolor{red}{DisCo~\cite{li2025disco}, a graph-based disentangled contrastive learning framework captures high-order user similarity relationships through multi-step random walks. scHNTL~\cite{meng2025schntl} identifies subtle cell similarities by constructing high-order adjacency matrices and optimizing cell embeddings using triplet loss.} HSDA~\cite{wang2025high} is introduced for speech emotion recognition, capturing high-order similarity through a similarity graph to reveal cross-domain structural relationships. Inspired by these methods and balance theory~\cite{cartwright1956structural}, this study propose the EDGL module, built upon the ADGL module, to fully exploit high-order similarity information for DTI prediction.

\section{Datasets and Methodology}

\textcolor{red}{In reality, drugs with similar structures may exhibit analogous mechanisms of action, while the similarity between targets reflects the functional conservation of targets~\cite{ding2014similarity}. Based on the observation that in human social networking, people engage in both direct friendships and higher-order connections through mutual friends, we draw an analogy in DTI prediction. We model the "direct friendships" by learning the known associations between drugs and targets. For the "higher-order friendships," we employ an even-polynomial graph filter based on balance theory. According to the triadic stability principle in balance theory, in a triadic relationship, if two relationships are positive, the third relationship tends to be positive as well~\cite{cartwright1956structural}. For instance, aspirin and salicylic acid share similar chemical structures and both exert their effects by inhibiting cyclooxygenase (COX)~\cite{feng2000prediction}. 
We hypothesize that, due to the underlying commonality between social interactions and drug-target relationships, strategies inspired by social friendship patterns can be effectively applied to DTI prediction~\cite{campillos2008drug}.}

\textcolor{red}{In this paper, we propose a dual-graph learning framework that simultaneously captures both direct similarities and higher-order indirect relationships within the heterogeneous drug-target graph, enhancing the accuracy of DTI prediction.}

\subsection{\textcolor{red}{Datasets and Preprocessing}}
To comprehensively assess the performance and applicability of SOC-DGL, this study selects four representative benchmark datasets: KIBA~\cite{tang2014making}, Davis~\cite{davis2011comprehensive}, BindingDB~\cite{liu2007bindingdb}, and DrugBank~\cite{wishart2018drugbank}, whose detailed description appears in Appendix Section A. \textcolor{red}{For each benchmark dataset, we extract various features of drugs and targets based on the SMILES sequences of drugs and the UniProt IDs of targets. For drug sequences, we use the RDKit software package to extract Molecular Access System (MACCS) keys fingerprints~\cite{durant2002reoptimization}, topological (TOP) fingerprints~\cite{carhart1985atom}, and Morgan fingerprints~\cite{morgan1965generation}, which represent three distinct aspects of the molecular features. For target sequences, to provide a more systematic description of target sequences and their physicochemical properties, we employ the iLearn software package to calculate amino acid composition (AAC), Moran autocorrelation (MORAN)~\cite{feng2000prediction}, Composition/Transition/Distribution (CTD)~\cite{cai2004enzyme}, and pseudo-amino acid composition (PAAC)~\cite{chou2005using}, which encompass four different aspects of target features.}

\subsection{Methodology}
\begin{figure*}
\centering
\includegraphics[width=\textwidth]{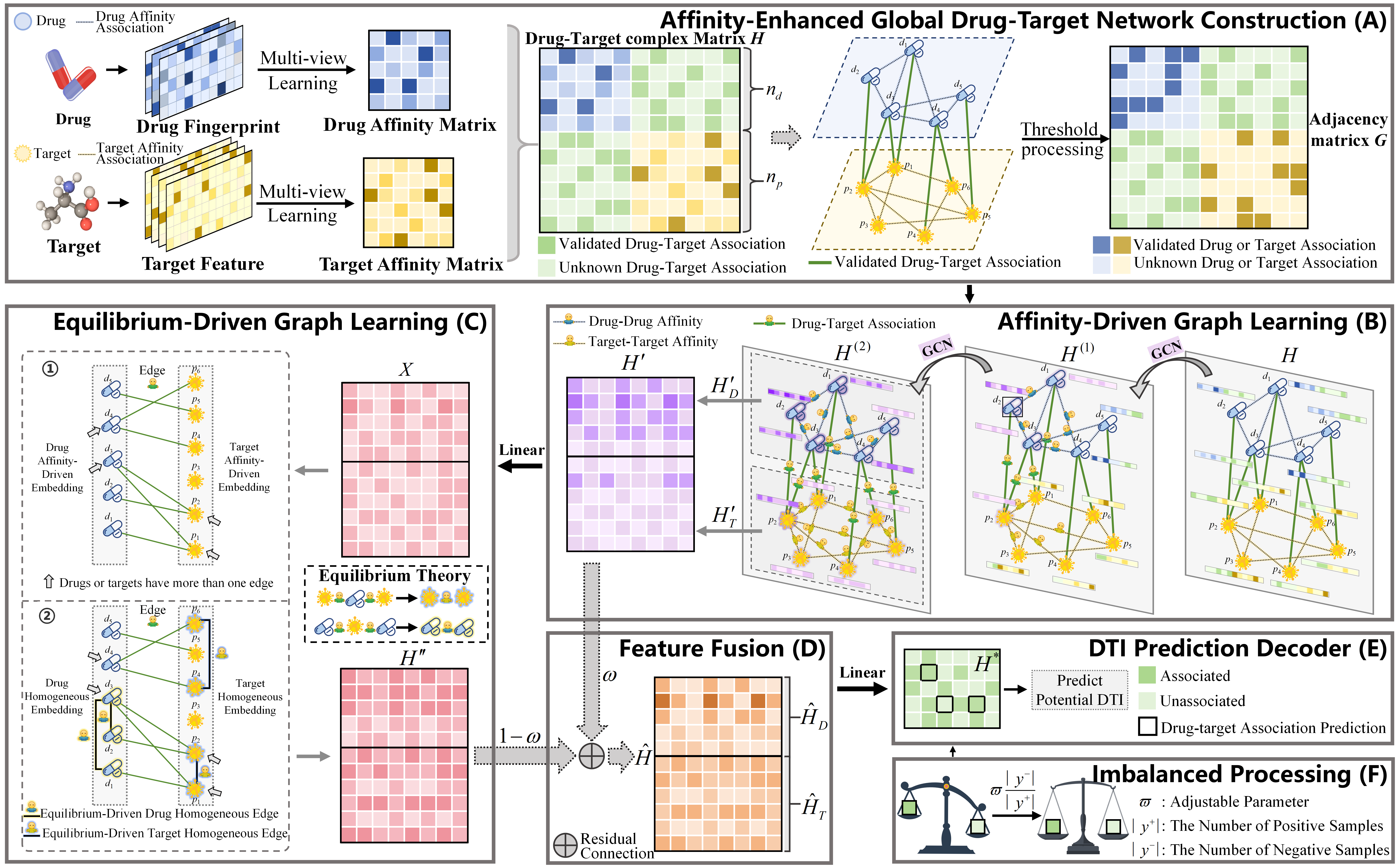}
\caption{The framework of SOC-DGL. (A) Affinity-enhanced global drug-target network construction, (B) Affinity-Driven Graph Learning, (C) Equilibrium-Driven Graph Learning, (D) Feature fusion, (E) DTI prediction decoder, and (F) Imbalanced Processing.}
\label{fig:1}
\end{figure*}

The framework of SOC-DGL is as shown in Fig.~\ref{fig:1}.

\subsubsection{Affinity-Enhanced Global Drug-Target Network Construction}
In DTI prediction, diverse features of drugs and targets are derived from various measurement methods, offering complementary perspectives on molecular characteristics. Integrating these features is crucial for accurate DTI inference. Assuming a shared latent space for drug and target features, SOC-DGL employs pairwise multi-view learning to derive a unified subspace representation from multiple views, specifically the drug affinity matrix (\( A_{DD} \)) and target affinity matrix (\( A_{TT} \)). SOC-DGL effectively integrates multi-view information by constructing a protocol matrix that combines correlation matrices under low-rank and sparsity constraints. \textcolor{red}{Details are provided in Appendix Section B (1).}

The drug-target affinity matrix is denoted as \( A_{DT} \). By combining \( A_{DD} \), \( A_{TT} \), and \( A_{DT} \), this study constructs an affinity-enhanced global drug-target network matrix \( H \), where \( A_{TD} \) is the transpose of \( A_{DT} \), and \( H \) is a symmetric matrix:

\begin{equation}
H = 
\begin{pmatrix}
A_{DD} & A_{DT} \\
A_{TD} & A_{TT}
\end{pmatrix} \in \mathbb{R}^{(n_d + n_t) \times (n_d + n_t)}
\label{eq:H_matrix}
\end{equation}
where \( n_d \) is the number of drugs, \( n_t \) is the number of targets.

\subsubsection{Affinity-Driven Graph Learning}


To effectively filter the useful information within the heterogeneous graph, for \( A_{DD} \) and \( A_{TT} \), elements in these two affinity matrices that are greater than or equal to the threshold 0.8 are set to 1, and the rest are set to 0, resulting in the drug affinity adjacency matrix \( \tilde{A}_{DD} \) and the target affinity adjacency matrix \( \tilde{A}_{TT} \). Therefore, the adjacency matrix \( G \) is constructed as:
\begin{equation}
G = 
\begin{pmatrix}
\tilde{A}_{DD} & A_{DT} \\
{A}_{TD} & \tilde{A}_{TT}
\end{pmatrix} \in \mathbb{R}^{(n_d+n_t) \times (n_d+n_t)}
\label{eq:H_prime}
\end{equation}

Then, this study performs graph convolution operations:
\begin{equation}
H^{(l+1)} = \sigma(D^{-\frac{1}{2}} G D^{-\frac{1}{2}} H^{(l)} W^{(l)})
\label{eq:H_prime}
\end{equation}
where \( D^{-1/2} G D^{-1/2} \) is the normalization process, with \( D = \mathrm{diag}\left(\sum_j G_{ij}\right) \) being the degree matrix of \( G \). \( H^{(l)} \) and \( W^{(l)} \) are the feature embedding matrix and learnable weight matrix at layer \( l \). \( \sigma \) is the activation function. After two rounds of graph convolution, the feature embedding matrix \( H' \) is obtained:
\begin{equation}
H' = 
\begin{bmatrix}
H'_D \\
H'_T
\end{bmatrix} \in \mathbb{R}^{(n_d + n_t) \times m}
\label{eq:H_prime}
\end{equation}
where \( H'_D \in \mathbb{R}^{n_d \times m} \) is the drug feature embedding matrix, \( H'_T \in \mathbb{R}^{n_t \times m} \) is the target feature embedding matrix, and \( m \) is the final output dimension.

\subsubsection{Equilibrium-Driven Graph Learning}


For ease of expression, let \( X_0 = H' \), and perform two linear transformations, applying an activation function. \( X \) is obtained as follows:
\begin{equation}
X_{(l+1)} = \sigma(X_{(l)} \cdot W_{(l)} + b_{(l)})
\label{eq:H_prime}
\end{equation}
This study normalizes the adjacency matrix \( G \) by setting \( G' = D^{-1/2} G D^{-1/2} \), where \( D \) is the degree matrix of the adjacency matrix \( G \). Thus,
\begin{equation}
G' = \begin{pmatrix}
\tilde{A}_{DD}' & A_{DT}' \\
A_{TD}' & \tilde{A}_{TT}'
\end{pmatrix} \in \mathbb{R}^{(n_d+n_t) \times (n_d+n_t)}
\label{eq:H_prime}
\end{equation}
To learn similarity information in drug-target association matrix, this study sets \textcolor{red}{\( \tilde{A}_{DD}' = 0 \), \( \tilde{A}_{TT}'= 0 \)}, thereby transforming it into the adjacency matrix \( A' \). Then, this study obtains the Laplacian matrix \( L = D' - A' \), where \( D' \) is the degree matrix of \( L \). \textcolor{red}{After standardizing \( L \),  \( \tilde{L} \) is obtained:}
\begin{equation}
\tilde{L} = D'^{-\frac{1}{2}} L D'^{-\frac{1}{2}}= I - D'^{-\frac{1}{2}} A' D'^{-\frac{1}{2}}
\label{eq:H_prime}
\end{equation}
where \( I \) is the identity matrix. Subsequently, this study defines the propagation matrix as \textcolor{red}{\( P = I - \tilde{L} = D’^{-1/2} A' D'^{-1/2} \)}, and the \( k \)-th order polynomial graph filter is defined as \( g(\tilde{L}) = \sum_{k=0}^{K} \alpha_k \tilde{L}^k \), \( k = 0, 1, \ldots, K \), where \( \alpha_k \) is the learnable weight matrix, and \( \alpha_k = \alpha \times (1 - \alpha)^k \), with \( \alpha \) being a hyperparameter that can be set. Thereafter, this study rewrites the graph filter formula as \( g(\tilde{L}) = \sum_{k=0}^{K} \alpha_k (I - \tilde{L})^k = \sum_{k=0}^{K} \alpha_k P^k \), discarding the odd-degree terms in \( g(\tilde{L}) \) and retaining only even-degree terms:
\begin{equation}
\textcolor{red}{H'' = \sum_{k=0}^{\left\lfloor \frac{K}{2} \right\rfloor} \alpha_{k} P^{2k}{X}}
\label{eq:H_prime}
\end{equation}
\textcolor{red}{where \( X \) is the input feature matrix that has already undergone two linear transformations.}


\subsubsection{Feature Fusion}
\textcolor{red}{This study employs residual connection techniques to integrate information from diverse sources, ensuring that the two modules complement each other and collectively form a more powerful graph reasoning mechanism. This approach balances the influence of higher-order interaction patterns on the generalized patterns, achieving equilibrium between the two social interaction paradigms. The corresponding formula is provided in Appendix Section B (2).}

\subsubsection{DTI Prediction Decoder}
DTI prediction decoder in SOC-DGL cleverly employs the principle of matrix factorization. By meticulously decoding these features, it reveals the affinity between drugs and targets, assessing the potential interaction affinity between different drugs and targets. This study denotes \( W^L \) as the learnable weight matrix for the mapping from the drug space to the target space within the decoder. Inspired by matrix tri-factorization, the formula is as follows:
\begin{equation}
H^* = \sigma(\hat{H}_D W^{(L)} \hat{H}_T^T)
\label{eq:H_prime}
\end{equation}
where \textcolor{red}{$\hat{H}_D$ and $\hat{H}_T$ are obtained through feature fusion}, \( H^* \) represents the final predicted interaction matrix. 

\subsubsection{Imbalanced Processing}
To address the issue of class imbalance between positive and negative samples, this study designs a Weighted Binary Cross-Entropy Loss Function with Reduced Regative Samples (RLF): 
\begin{equation}
\mathcal{L}=-\frac{1}{n_d\cdot n_t}(\varpi\frac{|y^-|}{|y^+|}\!\!\sum_{(i,j)\in y^+} \!\!\!\!\log(h_{ij}^*)+\!\!\!\!\sum_{(i,j)\in y^-}\!\!\!\!\log(1-h_{ij}^*))
\label{eq:H_prime}
\end{equation}
where \( |y^+| \) and \( |y^-| \) represent the number of positive samples and negative samples. \( h_{ij}^* \) represents the model’s predicted probability of interaction between drug \( i \) and target \( j \). This loss function is built upon the existing weighted cross-entropy loss function, with the introduction of an adjustable parameter $\varpi$, which effectively reduces the weights of negative samples while keeping the weights of positive samples unchanged.
\subsubsection{Parameter Setting}
\textcolor{red}{Experiments were conducted on four benchmark datasets under balanced or imbalanced setting. For the balanced KIBA dataset, we set the learning rate to 0.00003, neighbor aggregation depth $K$ to 200, neighbor contribution ratio $\alpha$ to 0.1, residual connection weight $w$ to 0.8, and maximum training epochs to 1500. Under the imbalanced setting, an additional imbalance loss parameter $\varpi$, used to adjust the weight of negative samples, was set to 0.6. Parameter details for other datasets under balanced or imbalanced setting are provided on our GitHub repository.}

\section{Experiments and Results}
\textcolor{red}{To evaluate the effectiveness of SOC-DGL, we design five experiments set, including the performance comparison, “Cold-start” Experiments, ablation experiments (consists of three sub-experiments), hyperparameter tuning experiments (consists of three sub-experiments) and case study.}

\subsection{Experimental Design Overview}
This study conducted validation experiments on both balanced (positive:negative = 1:1) and imbalanced (positive:negative = 1:10) datasets. Real-world tasks, particularly DTI prediction, typically involve imbalanced datasets with far fewer positive (known) than negative (unknown) samples, evaluating with imbalanced datasets provides a more realistic assessment of SOC-DGL's performance. Balancing the dataset, however, mitigates model bias towards predicting negative samples, offering a more balanced evaluation of model effectiveness. To minimize data variability, 10-fold cross-validation was employed to evaluate model performance. During each fold, the test set was masked, and the performance is evaluated using metrics such as AUROC, AUPR, F1\_score, ACC, Recall and Precision. For experiments, comprehensive validation was performed on four benchmark datasets to assess SOC-DGL’s performance. However, the KIBA dataset was primarily selected for other specific experiments due to its moderate size and comprehensive data, which ensured accurate model training and evaluation. This study conducted experiments using PyTorch 1.11.0 on an Nvidia GeForce RTX 3060 GPU, and employed the Xavier initialization strategy to prevent gradient explosion and vanishing during the training process.

\begin{table}
    \caption{\textcolor{red}{T-test Across Folds for SOC-DGL on Four Datasets under Balanced and Imbalanced Conditions.}}
    \begin{adjustbox}{width=0.487\textwidth}
    \tiny
    \begin{tabular}{@{}lcccc@{}}
        \toprule
        Dataset     & $P(\text{KIBA})$                             & $P(\text{Davis})$                             & $P(\text{BindingDB})$                         & $P(\text{DrugBank})$                             \\ \midrule
        Balanced  & \begin{tabular}[c]{@{}c@{}} 0.9950 \end{tabular} 
                           & \begin{tabular}[c]{@{}c@{}} 0.5959 \end{tabular}        & \begin{tabular}[c]{@{}c@{}} 0.2242 \end{tabular} 
                           & 0.5802                            \\ 
        Imbalanced  & \begin{tabular}[c]{@{}c@{}} 0.0133 \end{tabular} 
                           & \begin{tabular}[c]{@{}c@{}} 0.9292 \end{tabular}        & \begin{tabular}[c]{@{}c@{}} 0.4479 \end{tabular} 
                           & 0.0429                            \\ \bottomrule
    \end{tabular}
    \end{adjustbox}
    \label{tab:1.5}
\end{table}

\subsection{Performance Comparison}
\textcolor{red}{To assess SOC-DGL's effectiveness, comparative experiments were conducted with seven leading DTI prediction methods: GraphDTA~\cite{nguyen2021graphdta}, LRSpNM~\cite{wu2021novo}, MLMC~\cite{yan2022drug}, MULGA~\cite{ma2023mulga}, MSI-DTI~\cite{zhao2024msi}, BCMMDA~\cite{huang2024predicting} and MMDG-DTI~\cite{hua2025mmdg}, representing the state-of-the-art (SOTA). The Details of these methods were available in Appendix Section C.}

\begin{table*}[ht]
\centering
\caption{Performance Comparison of SOC-DGL with Baseline Methods on Four Datasets.}
\begin{tabular}{
@{}>{\raggedright\arraybackslash}p{1.3cm}|
>{\raggedright\arraybackslash}p{1.51cm}| 
>{\raggedright\arraybackslash}p{1.51cm}|  
>{\centering\arraybackslash}p{0.7cm}| 
>{\centering\arraybackslash}p{1.23cm}
>{\centering\arraybackslash}p{1.23cm}
>{\centering\arraybackslash}p{1.23cm}
>{\centering\arraybackslash}p{1.23cm}
>{\centering\arraybackslash}p{1.23cm}
>{\centering\arraybackslash}p{1.23cm}
>{\centering\arraybackslash}p{1.23cm}}
\toprule
\textbf{Dataset} & \textbf{Situation} & \textbf{Model} & \textbf{Year} & \textbf{AUROC} & \textbf{AUPR} & \textbf{F1\_score} & \textbf{ACC} & \textbf{Recall} & \textbf{Specificity} & \textbf{Precision} \\ \hline
\multirow{14}{*}{KIBA} & \multirow{7}{*}{Balanced} & GraphDTA & 2021 & 0.9295 & 0.9355 & 0.8783 & 0.8566 & 0.9237 & 0.7710 & 0.8372 \\ 
 &  & LRSpNM & 2021 & 0.9224 & 0.9382 & 0.8740 & 0.8740 & 0.8740 & 0.7730 & 0.8371 \\ 
 &  & MLMC & 2022 & 0.9484 & 0.9592 & 0.9049 & 0.8935 & 0.9038 & 0.8803 & 0.9059 \\ 
 &  & MULGA & 2023 & 0.9479 & 0.9551 & 0.8984 & 0.8773 & 0.9404 & 0.7911 & 0.8600 \\
 &  & \textcolor{red}{MSI-DTI} & \textcolor{red}{2024} & \textcolor{red}{0.9320} & \textcolor{red}{0.8370} & \textcolor{red}{0.7503} & \textcolor{red}{0.9080} & \textcolor{red}{0.7210} & \textcolor{red}{-} & \textcolor{red}{0.7820} \\
 &  & MMDG-DTI & 2025 & 0.9586 & 0.9646 & 0.9083 & 0.8949 & 0.9318 & 0.8483 & 0.8858 \\ 
 &  & \cellcolor{gray!17}SOC-DGL & \cellcolor{gray!17}\textbackslash & \cellcolor{gray!17}\textbf{0.9761} & \cellcolor{gray!17}\textbf{0.9805} & \cellcolor{gray!17}\textbf{0.9350} & \cellcolor{gray!17}\textbf{0.9268} & \cellcolor{gray!17}\textbf{0.9373} & \cellcolor{gray!17}\textbf{0.9133} & \cellcolor{gray!17}\textbf{0.9328} \\ \cline{2-11} 
 & \multirow{7}{*}{Imbalanced} & GraphDTA & 2021 & 0.9195 & 0.6500 & 0.6121 & 0.8911 & 0.6895 & 0.9198 & 0.5502 \\ 
 &  & LRSpNM & 2021 & 0.9132 & 0.6904 & 0.6339 & 0.8998 & 0.6900 & 0.9299 & 0.5863 \\ 
 &  & MLMC & 2022 & 0.9329 & 0.7462 & 0.6874 & 0.9192 & 0.7084 & 0.9494 & 0.6675 \\ 
 &  & MULGA & 2023 & 0.8918 & 0.5556 & 0.5628 & 0.8666 & 0.6808 & 0.8934 & 0.4796 \\ 
 &  & \textcolor{red}{BCMMDA} & \textcolor{red}{2024} & \textcolor{red}{0.9270} & \textcolor{red}{0.8040} & \textcolor{red}{-} & \textcolor{red}{-} & \textcolor{red}{-} & \textcolor{red}{-} & \textcolor{red}{-} \\
 &  & MMDG-DTI & 2025 & 0.9245 & 0.7193 & 0.6639 & 0.9102 & 0.6998 & 0.9391 & 0.6273 \\ 
 \rowcolor{gray!17}
\cellcolor{white}& \cellcolor{white}  & SOC-DGL & \textbackslash & \textbf{0.9364} & \textbf{0.7473} & \textbf{0.6908} & \textbf{0.9207} & \textbf{0.7103} & \textbf{0.9506} & \textbf{0.6730} \\ \hline
\multirow{14}{*}{Davis} & \multirow{7}{*}{Balanced} & GraphDTA & 2021 & 0.9015 & 0.9324 & 0.8707 & 0.8330 & 0.9016 & 0.7195 & 0.8418 \\ 
 &  & LRSpNM & 2021 & 0.8930 & 0.9256 & 0.8644 & 0.8255 & 0.9083 & 0.6951 & 0.8250 \\ 
 &  & MLMC & 2022 & 0.9072 & 0.9395 & 0.8712 & 0.8712 & \textbf{0.9098} & 0.7177 & 0.8356 \\ 
 &  & MULGA & 2023 & 0.9158 & 0.9451 & 0.8790 & 0.8561 & 0.8539 & \textbf{0.8596} & \textbf{0.9058} \\ 
 &  & \textcolor{red}{MSI-DTI} & \textcolor{red}{2024} & \textcolor{red}{\textbf{0.9660}} & \textcolor{red}{0.8930} & \textcolor{red}{0.8570} & \textcolor{red}{\textbf{0.9325}} &  \textcolor{red}{0.8805} & \textcolor{red}{-} & \textcolor{red}{0.8348} \\
 &  & MMDG-DTI & 2025 & 0.9356 & 0.9647 & 0.8905 & 0.8627 & 0.9057 & 0.7934 & 0.8758 \\ 
 &  & \cellcolor{gray!17}SOC-DGL & \cellcolor{gray!17}\textbackslash & \cellcolor{gray!17}0.9467 & \cellcolor{gray!17}\textbf{0.9673} & \cellcolor{gray!17}\textbf{0.9038} & \cellcolor{gray!17}0.8816 & \cellcolor{gray!17}0.9097 & \cellcolor{gray!17}0.8372 & \cellcolor{gray!17}0.8985 \\ \cline{2-11} 
 & \multirow{7}{*}{Imbalanced} & GraphDTA & 2021 & 0.8486 & 0.6024 & 0.5514 & 0.8797 & 0.5096 & 0.9425 & 0.6006 \\ 
 &  & LRSpNM & 2021 & 0.9108 & 0.7067 & 0.6415 & 0.8961 & 0.6348 & 0.9409 & 0.6517 \\ 
 &  & MLMC & 2022 & 0.9110 & 0.7253 & 0.6704 & 0.9057 & 0.6530 & 0.9492 & 0.6888 \\ 
 &  & MULGA & 2023 & 0.8812 & 0.6179 & 0.5710 & 0.8698 & 0.5874 & 0.9187 & 0.5556 \\ 
 &  & \textcolor{red}{BCMMDA} & \textcolor{red}{2024} & \textcolor{red}{\textbf{0.9330}} & \textcolor{red}{\textbf{0.7800}} & \textcolor{red}{0.7040} & \textcolor{red}{0.8760} & \textcolor{red}{0.6070} & \textcolor{red}{-} & \textcolor{red}{0.8380} \\
 &  & MMDG-DTI & 2025 & 0.9087 & 0.7787 & \textbf{0.7369} & \textbf{0.9326} & 0.6503 & \textbf{0.9805} & \textbf{0.8409} \\ 
 &  & \cellcolor{gray!17}SOC-DGL & \cellcolor{gray!17}\textbackslash & \cellcolor{gray!17}0.9192 & \cellcolor{gray!17}0.7655 \cellcolor{gray!17}& \cellcolor{gray!17}0.6918 & \cellcolor{gray!17}0.9134 & \cellcolor{gray!17}\textbf{0.6642} & \cellcolor{gray!17}0.9561 & \cellcolor{gray!17}0.7238 \\ \hline
\multirow{12}{*}{BindingDB} & \multirow{6}{*}{Balanced} & GraphDTA & 2021 & 0.9325 & 0.9271 & 0.8719 & 0.8605 & 0.9103 & 0.8063 & 0.8366 \\ 
 &  & LRSpNM & 2021 & 0.8514 & 0.5771 & 0.5533 & 0.8706 & 0.5464 & 0.9263 & 0.5602 \\ 
 &  & MLMC & 2022 & 0.8797 & 0.9285 & 0.8707 & 0.8750 & 0.8237 & 0.9286 & 0.9233 \\ 
 &  & MULGA & 2023 & 0.9874 & 0.9859 & \textbf{0.9678} & \textbf{0.9660} & 0.9818 & 0.9490 & 0.9542 \\ 
 &  & MMDG-DTI & 2025 & 0.9620 & 0.9661 & 0.9181 & 0.9172 & 0.9172 & 0.9340 & 0.9340 \\ 
 &  & \cellcolor{gray!17}SOC-DGL & \cellcolor{gray!17}\textbackslash & \cellcolor{gray!17}\textbf{0.9894} & \cellcolor{gray!17}\textbf{0.9888} & \cellcolor{gray!17}0.9641 & \cellcolor{gray!17}0.9627 & \cellcolor{gray!17}\textbf{0.9677} & \cellcolor{gray!17}\textbf{0.9574} & \cellcolor{gray!17}\textbf{0.9606} \\ \cline{2-11} 
 & \multirow{6}{*}{Imbalanced} & GraphDTA & 2021 & 0.9718 & 0.8099 & 0.7614 & 0.9320 & 0.7724 & 0.9718 & 0.7508 \\ 
 &  & LRSpNM & 2021 & 0.8481 & 0.5727 & 0.5866 & 0.9096 & 0.6383 & 0.9399 & 0.5426 \\ 
 &  & MLMC & 2022 & 0.8992 & 0.8260 & 0.7990 & \textbf{0.9617} & 0.7584 & \textbf{0.9844} & \textbf{0.8443} \\ 
 &  & MULGA & 2023 & 0.9706 & 0.8366 & 0.7188 & 0.9444 & 0.6991 & 0.9721 & 0.7395 \\ 
 &  & MMDG-DTI & 2025 & 0.9149 & 0.7428 & 0.6967 & 0.8980 & 0.6504 & 0.9508 & 0.7429 \\ 
 &  & \cellcolor{gray!17}SOC-DGL & \cellcolor{gray!17}\textbackslash & \cellcolor{gray!17}\textbf{0.9790} & \cellcolor{gray!17}\textbf{0.8705} & \cellcolor{gray!17}\textbf{0.8057} & \cellcolor{gray!17}0.9594 & \cellcolor{gray!17}\textbf{0.8406} & \cellcolor{gray!17}0.9726 & \cellcolor{gray!17}0.7744 \\ \hline
\multirow{14}{*}{DrugBank} & \multirow{7}{*}{Balanced} & GraphDTA & 2021 & 0.8762 & 0.8676 & 0.8304 & 0.8033 & \textbf{0.9242} & 0.6719 & 0.7539 \\ 
 &  & LRSpNM & 2021 & 0.7563 & 0.8206 & 0.7016 & 0.6687 & 0.7522 & 0.5790 & 0.6573 \\ 
 &  & MLMC & 2022 & 0.8637 & 0.9126 & 0.8371 & 0.8415 & 0.7828 & 0.9052 & 0.8995 \\ 
 &  & MULGA & 2023 & 0.9715 & 0.9743 & 0.9165 & 0.9141 & 0.9112 & 0.9171 & 0.9219 \\ 
 &  & \textcolor{red}{MSI-DTI} & \textcolor{red}{2024} & \textcolor{red}{0.9093} & \textcolor{red}{0.9136} & \textcolor{red}{0.8339} & \textcolor{red}{0.8381} & \textcolor{red}{0.8551} & \textcolor{red}{-} & \textcolor{red}{0.8137} \\
 &  & MMDG-DTI & 2025 & 0.9626 & 0.9658 & 0.9180 & 0.9159 & 0.9083 & 0.9240 & 0.9281 \\ 
 &  & \cellcolor{gray!17}SOC-DGL & \cellcolor{gray!17}\textbackslash & \cellcolor{gray!17}\textbf{0.9735} & \cellcolor{gray!17}\textbf{0.9774} & \cellcolor{gray!17}\textbf{0.9286} & \cellcolor{gray!17}\textbf{0.9267} & \cellcolor{gray!17}0.9208 & \cellcolor{gray!17}\textbf{0.9331} & \cellcolor{gray!17}\textbf{0.9369} \\ \cline{2-11} 
 & \multirow{7}{*}{Imbalanced} & GraphDTA & 2021 & 0.8450 & 0.4268 & 0.4564 & 0.8764 & 0.5336 & 0.9133 & 0.3987 \\ 
 &  & LRSpNM & 2021 & 0.7752 & 0.5480 & 0.5437 & 0.9257 & 0.4577 & 0.9758 & 0.6695 \\ 
 &  & MLMC & 2022 & 0.8594 & 0.7330 & 0.7043 & 0.9498 & 0.6181 & \textbf{0.9853} & \textbf{0.8185} \\ 
 &  & MULGA & 2023 & 0.9259 & 0.7179 & 0.6667 & 0.9358 & 0.6579 & 0.9659 & 0.6756 \\ 
 &  & \textcolor{red}{BCMMDA} & \textcolor{red}{2024} & \textcolor{red}{0.8560} & \textcolor{red}{\textbf{0.8680}} & \textcolor{red}{\textbf{0.7895}} & \textcolor{red}{0.7860} & \textcolor{red}{\textbf{0.7920}} & \textcolor{red}{-} & \textcolor{red}{0.7870} \\
 &  & MMDG-DTI & 2025 & 0.9051 & 0.6978 & 0.6717 & 0.9381 & 0.6522 & 0.9688 & 0.6943 \\ 
 & & \cellcolor{gray!17}SOC-DGL & \cellcolor{gray!17}\textbackslash & \cellcolor{gray!17}\textbf{0.9295} & \cellcolor{gray!17}0.7654 & \cellcolor{gray!17}0.7211 & \cellcolor{gray!17}\textbf{0.9508} & \cellcolor{gray!17}0.6567 & \cellcolor{gray!17}0.9824 & \cellcolor{gray!17}0.8004 \\ \toprule
\end{tabular}
\label{tab:1}
\end{table*}

\textcolor{red}{To validate the model's robustness, this study conducted ten-fold cross-validation of SOC-DGL on four balanced and two imbalanced datasets. Paired t-tests were performed on seven metrics for each fold, followed by Fisher's Combined Probability Test to aggregate the results into a single p-value. As shown in Table~\ref{tab:1.5}, the combined p-values for SOC-DGL on the four balanced and two imbalanced datasets are all greater than 0.05, except for the KIBA and DrugBank imbalanced datasets. This suggests that SOC-DGL is stable on balanced datasets and certain imbalanced datasets, demonstrating robustness, likely due to its adjustable imbalance loss function.}

\textcolor{red}{As shown in Table~\ref{tab:1}, the results demonstrate that the SOC-DGL model achieves robust overall performance across the four benchmark datasets. Specifically, except for the Davis dataset under the highly imbalanced data scenario where SOC-DGL performs slightly worse than MCMMDA, the model attains or surpasses SOTA results on the other datasets. This study speculate that this discrepancy primarily arises from the Davis dataset’s pronounced class imbalance, where the number of positive samples is significantly lower than negatives, posing challenges for the model to effectively learn minority class features. Nevertheless, SOC-DGL’s superior performance across diverse and relatively balanced datasets highlights the effectiveness of its graph-structured learning approach in capturing complex drug-target relationships. These results validate the soundness and practical potential of SOC-DGL.}

\subsection{“Cold-start” Experiments}
In the “cold-start” experiments, two scenarios were designed: “cold-start for drugs” and “cold-start for targets,” where the drug-target affinity matrix was systematically sampled. In the “cold-start for targets” scenario, drugs linked to specific targets were removed, treating each target as new, with the removed data serving as the test set. Similar operations were applied in the "cold-start for drugs" scenario.

The KIBA dataset contains 1,720 drugs and 220 targets, providing the foundation for the "cold-start" experiments. \textcolor{red}{A total of 1940 independent cold-start experiments were conducted, including 1720 experiments for drugs and 220 experiments for targets. These experiments were carried out by using information from other known drugs and targets to predict the selected drugs and targets, without employing 10-fold cross-validation.} This setup simulates real-world scenarios where new drugs or targets lack historical interaction data, which is especially relevant in drug discovery, where novel compounds or under-explored proteins often lack prior interaction data.

\begin{figure}[H]
\centering
\includegraphics[width=0.495\textwidth]{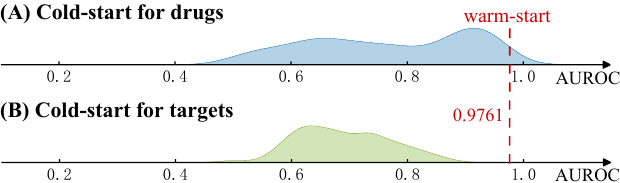}
\caption{\textcolor{red}{The performance of SOC-DGL in "cold-start" scenarios for drugs or targets, where the peak represents the degree of aggregation of results.}}
\label{fig:3}
\end{figure}

\begin{figure*}[t]
\centering
\includegraphics[width=\textwidth]{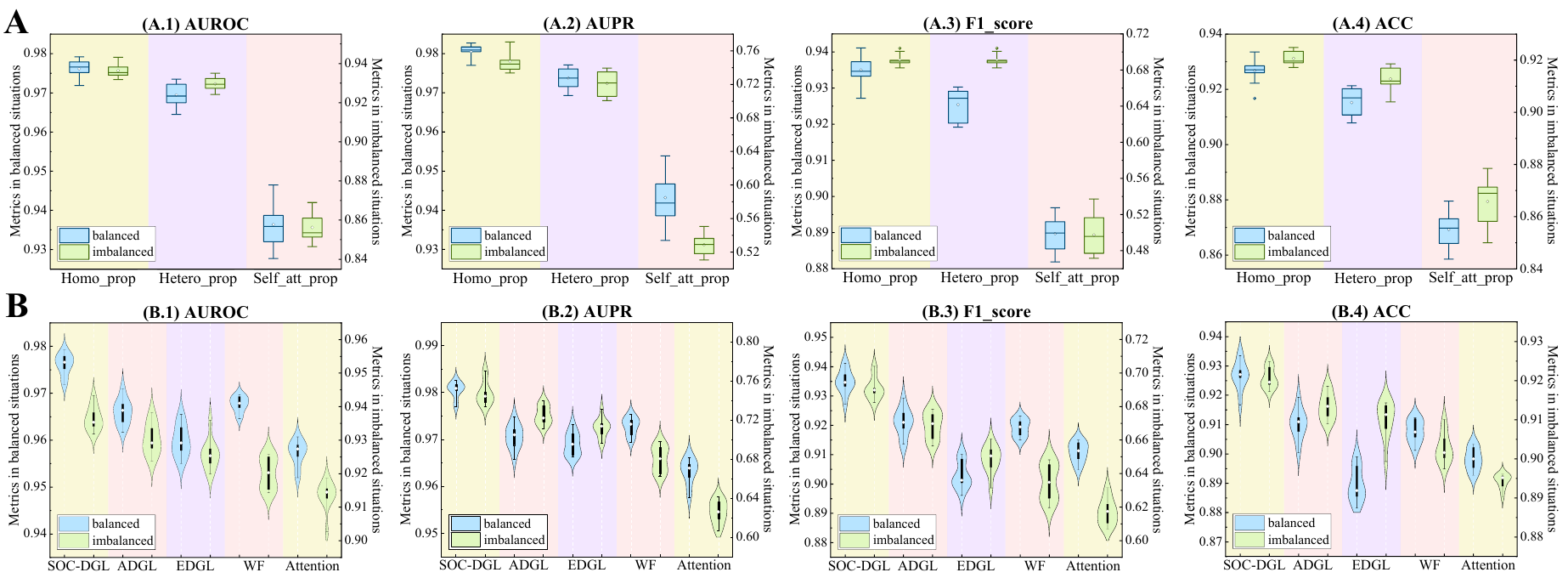}
\caption{Ablation Experiments on the KIBA dataset. (A) Experimental evaluation of Homo\_prop, Hetero\_prop, and Self\_att\_prop. 
(B) \textcolor{red}{Ablation experiment on ADGL, EDGL, and feature fusion strategies (WF and Attention) for model performance.} }
\label{fig:4}
\end{figure*}

As shown in Fig.~\ref{fig:3}, \textcolor{red}{in the "cold-start for drugs" scenario. SOC-DGL demonstrated stable performance with AUROC values ranging from [0.6, 1.0], with the most frequent results around 0.9, indicating its strong generalization ability for new drugs. In the "cold-start for targets" scenario, the AUROC values ranged from [0.6, 0.8], which may have been influenced by the availability of target data and variability in target-specific interactions. These results confirmed SOC-DGL's robustness in predicting DTI for unseen drugs or targets, making it a valuable tool for practical drug discovery.}


\begin{table*}[htbp]
\centering
\caption{Warm start validation experiment on four benchmark imbalanced datasets using different loss functions.}
\setlength{\tabcolsep}{3pt}
\scriptsize
\begin{tabular}{
>{\raggedright\arraybackslash}p{2cm}|
>{\centering\arraybackslash}p{1.25cm}| 
>{\centering\arraybackslash}p{1.85cm} 
>{\centering\arraybackslash}p{1.85cm}
>{\centering\arraybackslash}p{1.85cm}
>{\centering\arraybackslash}p{1.85cm}
>{\centering\arraybackslash}p{1.85cm}
>{\centering\arraybackslash}p{1.85cm}
>{\centering\arraybackslash}p{1.85cm}}
\toprule
\textbf{Dataset} & \textbf{Loss} & \textbf{AUROC} & \textbf{AUPR} & \textbf{F1\_score} & \textbf{ACC} & \textbf{Recall} & \textbf{Specificity} & \textbf{Precision} \\
\midrule
\multirow{4}{*}{KIBA dataset} 
& SLF & 0.9329 $\pm$ 0.0042 & 0.7300 $\pm$ 0.0111 & 0.6758 $\pm$ 0.0090 & 0.9156 $\pm$ 0.0036  & 0.7054 $\pm$ 0.0265 & 0.9455 $\pm$ 0.0071 & 0.6501 $\pm$ 0.0230 \\
& FLF & 0.8983 $\pm$ 0.0078 & 0.6209 $\pm$ 0.0029 & 0.5400 $\pm$ 0.0023 & 0.8653 $\pm$ 0.0145 & 0.6530 $\pm$ 0.0175 & 0.8130 $\pm$ 0.0187 & 0.5411 $\pm$ 0.0016 \\
& WLF & 0.9309 $\pm$ 0.0035 & 0.7081 $\pm$ 0.0074 & 0.6675 $\pm$ 0.0056 & 0.9126 $\pm$ 0.0020 & 0.7032 $\pm$ 0.0263 & 0.9425 $\pm$ 0.0058 & 0.6363 $\pm$ 0.0164 \\
\rowcolor{gray!17}
\cellcolor{white}
& \textbf{RLF} & \textbf{0.9364 $\pm$ 0.0036 } & \textbf{0.7473 $\pm$ 0.0123 } & \textbf{0.6908 $\pm$ 0.0068 } & \textbf{0.9207  $\pm$ 0.0026} &\textbf{0.7103 $\pm$ 0.0182} & \textbf{0.9506 $\pm$ 0.0049 } & \textbf{0.6730 $\pm$ 0.0177 } \\
\midrule
\multirow{4}{*}{Davis dataset} 
& SLF & 0.9171  $\pm$ 0.0046 & 0.7601  $\pm$ 0.0089 & 0.6907  $\pm$ 0.0144 & \textbf{0.9135 $\pm$ 0.0061 } & 0.6618  $\pm$ 0.0234 & 0.9558  $\pm$ 0.0085 & 0.7233  $\pm$ 0.0334 \\
& FLF & 0.8351 $\pm$ 0.0106  & 0.6934 $\pm$ 0.0094  & 0.6132 $\pm$ 0.0122  & 0.8345 $\pm$ 0.0199  & 0.5938 $\pm$ 0.0443  & 0.8687 $\pm$ 0.0292  & 0.6376 $\pm$ 0.0161  \\
& WLF & 0.9158 $\pm$ 0.0031 & 0.7506 $\pm$ 0.0085 & 0.6790 $\pm$ 0.0038 & 0.9097 $\pm$ 0.0042 & 0.6516 $\pm$ 0.0221 & 0.9540 $\pm$ 0.0082 & 0.7102 $\pm$ 0.0592 \\
\rowcolor{gray!17}
\cellcolor{white}
& \textbf{RLF} & \textbf{0.9192 $\pm$ 0.0045} & \textbf{0.7655 $\pm$ 0.0134} & \textbf{0.6918 $\pm$ 0.0129} & 0.9134  $\pm$ 0.0030 & \textbf{0.6642 $\pm$ 0.0338} & \textbf{0.9561 $\pm$ 0.0066} & \textbf{0.7238 $\pm$ 0.0230} \\
\midrule
\multirow{4}{*}{BindingDB dataset} 
& SLF & 0.9684 $\pm$ 0.0064 & 0.8031 $\pm$ 0.0302 & 0.7553 $\pm$ 0.0350 & 0.9467 $\pm$ 0.0101 & 0.8162 $\pm$ 0.0311 & 0.9612 $\pm$ 0.0116 & 0.7060 $\pm$ 0.0036 \\
& FLF & 0.9067 $\pm$ 0.0151 & 0.7535 $\pm$ 0.0167 & 0.7807 $\pm$ 0.0151 & 0.8996 $\pm$ 0.0109 & 0.8139 $\pm$ 0.0140 & 0.9003 $\pm$ 0.0128 & 0.6994 $\pm$ 0.0268 \\
& WLF & 0.9578 $\pm$ 0.0074  & 0.7800 $\pm$ 0.0282  & 0.6899 $\pm$ 0.0405  & 0.9336 $\pm$ 0.0116  & 0.7326 $\pm$ 0.0392  & 0.9560 $\pm$ 0.0134  & 0.6564 $\pm$ 0.0685  \\
\rowcolor{gray!17}
\cellcolor{white}
& \textbf{RLF} & \textbf{0.9790 $\pm$ 0.0034 } & \textbf{0.8705 $\pm$ 0.0146 } & \textbf{0.8057 $\pm$ 0.0177 } & \textbf{0.9594 $\pm$ 0.0041 } & \textbf{0.8406 $\pm$ 0.0246}  & \textbf{0.9726 $\pm$ 0.0047 } & \textbf{0.7744 $\pm$ 0.0304 } \\
\midrule
\multirow{4}{*}{DrugBank dataset} 
& SLF & 0.9150 $\pm$ 0.0094 & 0.7316 $\pm$ 0.0256 & 0.6998 $\pm$ 0.0205 & 0.9486 $\pm$ 0.0034 & 0.6158 $\pm$ 0.0267 & \textbf{0.9845 $\pm$ 0.0022} & \textbf{0.8104 $\pm$ 0.0206} \\
& FLF & 0.8206 $\pm$ 0.0096 & 0.4512 $\pm$ 0.0077 & 0.5127 $\pm$ 0.0133 & 0.8756 $\pm$ 0.0321 & 0.4733 $\pm$ 0.0331 & 0.9511 $\pm$ 0.0388 & 0.5755 $\pm$ 0.0414 \\
& WLF & 0.9208 $\pm$ 0.0073 & 0.7321 $\pm$ 0.0204 & 0.6795 $\pm$ 0.0164 & 0.9388 $\pm$ 0.0029  & 0.6696 $\pm$ 0.0242  & 0.8677 $\pm$ 0.0030  & 0.6897 $\pm$ 0.0249 \\
\rowcolor{gray!17}
\cellcolor{white}
& \textbf{RLF} & \textbf{0.9295 $\pm$ 0.0076 } & \textbf{0.7654 $\pm$ 0.0126 } & \textbf{0.7211 $\pm$ 0.0104 } & \textbf{0.9508 $\pm$ 0.0015 } & \textbf{0.6567 $\pm$ 0.0210 } & 0.9824 $\pm$ 0.0022  & 0.8004 $\pm$ 0.0174  \\
\bottomrule
\end{tabular}
\label{tab:2}
\end{table*}

\subsection{Ablation Experiments}
\subsubsection{Ablation Experiment on Even-Polynomial Graph Filter in EDGL}
To analyze the importance of the even-polynomial graph filter in EDGL for indirectly extracting higher-order homogeneous information between drugs and targets, this study compared it with the odd-polynomial graph filter for extracting higher-order heterogeneous information and the self-attention mechanism for enhancing self-information under the warm-start scenario, while keeping all other modules unchanged. \textcolor{red}{Specifically, this study evaluated the performance of three variants: Homo\_prop (even-polynomial filter for homogeneous similarity enhancement), Hetero\_prop (odd-polynomial filter for heterogeneous interaction enhancement), and Self\_att\_prop (self-attention mechanism for self-information enhancement).} As shown in Fig.~\ref{fig:4} (A), Homo\_prop consistently achieved superior performance across all evaluation metrics. These results demonstrate the effectiveness of the even-polynomial graph filter in capturing complex high-order homogeneous relationships, enabling more accurate and robust DTI prediction.

\subsubsection{Ablation Experiment on ADGL, EDGL, and Feature Fusion Strategies to Model Performance}

\textcolor{red}{To evaluate the impact of different graph learning modules, we tested four SOC-DGL variants: (1) ADGL, focusing on global associations, (2) EDGL, emphasizing higher-order similarities, (3) WF, where features pass through ADGL and EDGL sequentially without fusion, and (4) Attention, fusing features from ADGL and EDGL via an attention mechanism after concatenation. As shown in Fig.~\ref{fig:4} (B), the ADGL module significantly contributes to model performance due to its extensive use of a global drug-target affinity network that captures both global DTI and individual similarity information. In contrast, the EDGL module amplifies the influence of even-hop neighbors through a higher-order strategy, enhancing structural information. Regarding feature fusion, the residual connection consistently outperforms other methods in predictive performance, offering computational advantages by avoiding matrix multiplication, which reduces parameters and training time. Biologically, the residual connection preserves the integrity of global drug-target associations and higher-order structural similarities, aligning with how functional and structural signals contribute to molecular recognition without excessive entanglement. In contrast, concatenation and attention mechanisms may obscure this interpretability by over-mixing heterogeneous information.}

\subsubsection{Ablation Experiment on Imbalanced Loss Function}
In DTI prediction, the imbalance between positive and negative samples can affect model performance, causing it to favor the majority class. This experiment evaluated the impact of various loss functions on model performance, focusing on those designed to address class imbalance: Standard Binary Cross-Entropy Loss (SLF), Focal Loss with Dynamic Scaling (FLF), Weighted Binary Cross-Entropy Loss (WLF), and Weighted Binary Cross-Entropy Loss with Reduced Negative Samples (RLF). Experiments were conducted in imbalanced datasets and details of these loss functions could be found in Appendix Section E.
As shown in Table~\ref{tab:2}, the RLF loss function consistently outperformed the others across four benchmark imbalanced datasets, demonstrating its effectiveness for imbalanced DTI prediction. Although RLF did not always yield the highest scores on individual metrics, it showed stable and robust performance. By incorporating a weighting parameter \( \varpi \) to reduce the weight of negative samples while maintaining the positive sample weight, RLF addressed class imbalance while preserving the model's discriminative ability. These results, observed across four distinct imbalanced datasets, further validated RLF's strong generalization capability.

\begin{figure*}[t]
\centering
\includegraphics[width=\textwidth]{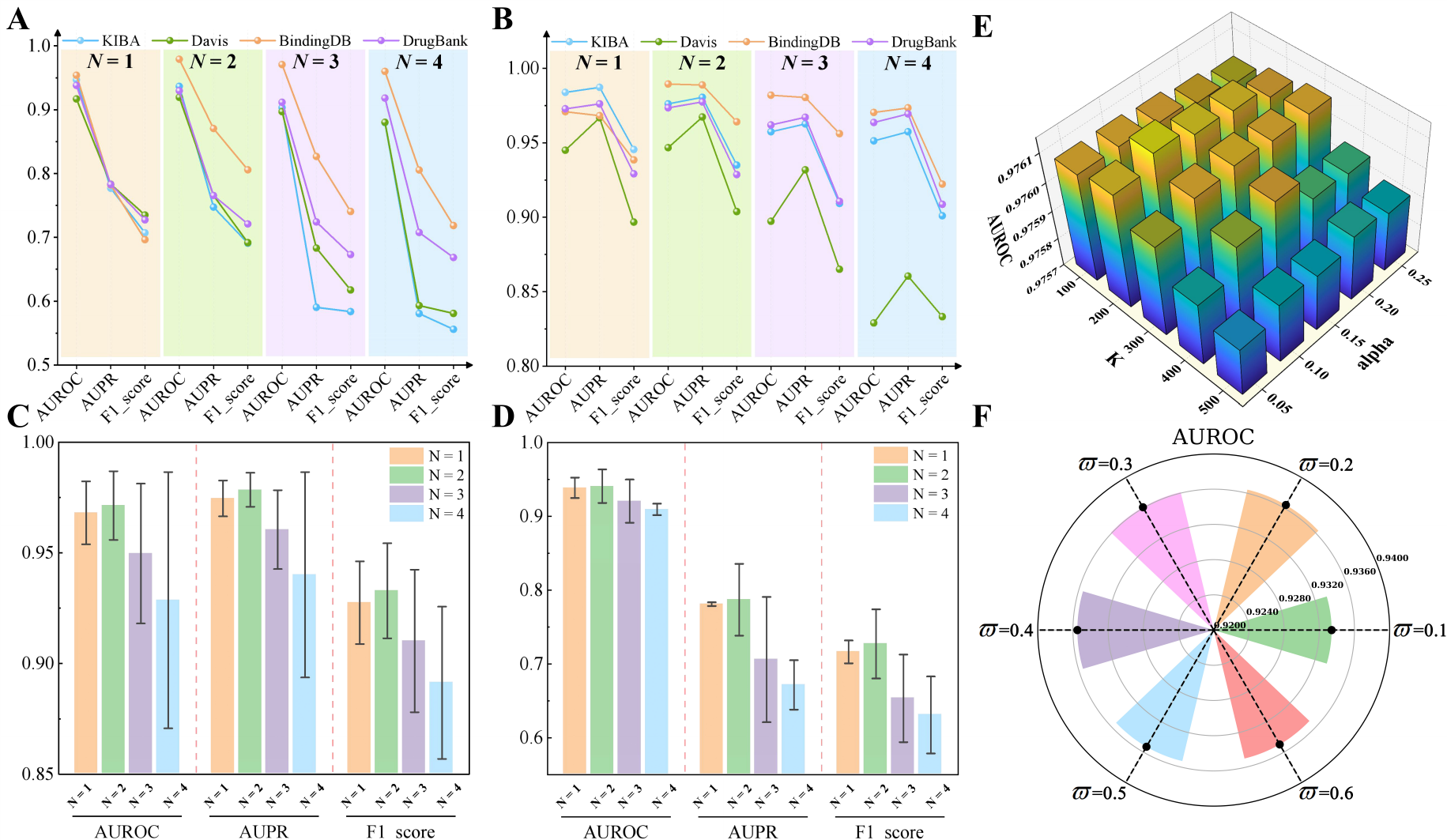}
\caption{SOC-DGL's performance across AUROC, AUPR, and F1\_score with different GCN layer settings on (A) four balanced datasets and (B) four imbalanced datasets. AUROC, AUPR, and F1\_score of SOC-DGL under different GCN layer settings on (C) balanced datasets and (D) imbalanced datasets. (E) Hyperparameter tuning of $K$ and $\alpha$ on the KIBA balanced dataset. (E) Hyperparameter tuning of $\varpi$ on the KIBA imbalanced dataset.}
\label{fig:5}
\end{figure*}

\subsection{Hyperparameter Tuning Experiments}
\textcolor{red}{To assess the model's robustness across multiple benchmark datasets, this study focused on three kinds of hyperparameters in SOC-DGL and systematically optimized these parameters.}

\subsubsection{Optimization of ADGL Hyperparameters}
The number of convolutional layers ($N$) in the ADGL module significantly influences the ability to capture the topological structure information. To evaluate the impact of layer configurations on DTI prediction, experiments were conducted on both balanced and imbalanced versions of four benchmark datasets, as shown in Fig.~\ref{fig:5} (A-D). Results demonstrated that a two-layer GCN consistently yielded the best performance. However, on the KIBA dataset, $N=1$ slightly outperformed $N=2$ in AUROC and AUPR, indicating that a shallower GCN better captured global associations in this specific case. In general, the two-layer configuration expanded the receptive field, enhancing message passing and model expressiveness, while additional layers increased complexity, raising the risk of overfitting.

\subsubsection{Optimization of EDGL Hyperparameters}
This study focused on optimizing key hyperparameters within the EDGL module, specifically the total number of iterations ($K$) and the influence parameter ($\alpha$), using the KIBA dataset. A range of values for $K$ (100 to 500) and $\alpha$ (0.05 to 0.25) were tested to evaluate their impact on SOC-DGL performance. As shown in Fig.~\ref{fig:5} (E), optimal performance was achieved with $K = 200$ and $\alpha = 0.20$, resulting in an AUROC of 0.9761. These results indicate that carefully balancing $K$ and $\alpha$ enhances model performance, while extreme values adversely affect results.

\subsubsection{Optimization of Adjustable Parameter $\varpi$}
Traditional imbalanced loss functions address the imbalance by adjusting the influence of positive samples based on their ratio to negative samples. However, it can lead to an excessive influence of positive sample loss, creating artificial positive sample imbalance. To mitigate this, an adjustable parameter $\varpi$ was introduced to fine-tune the correction for positive samples. Hyperparameter experiments with $\varpi \in \{0.1, 0.2, 0.3, 0.4, 0.5, 0.6\}$ were conducted, and the results in Fig.~\ref{fig:5} (F) demonstrated that $\varpi = 0.2$ achieved optimal performance. These findings emphasize the importance of appropriately tuning the adjustable parameter to effectively handle sample imbalance.


\begin{figure*}[t]
\centering
\includegraphics[width=\textwidth]{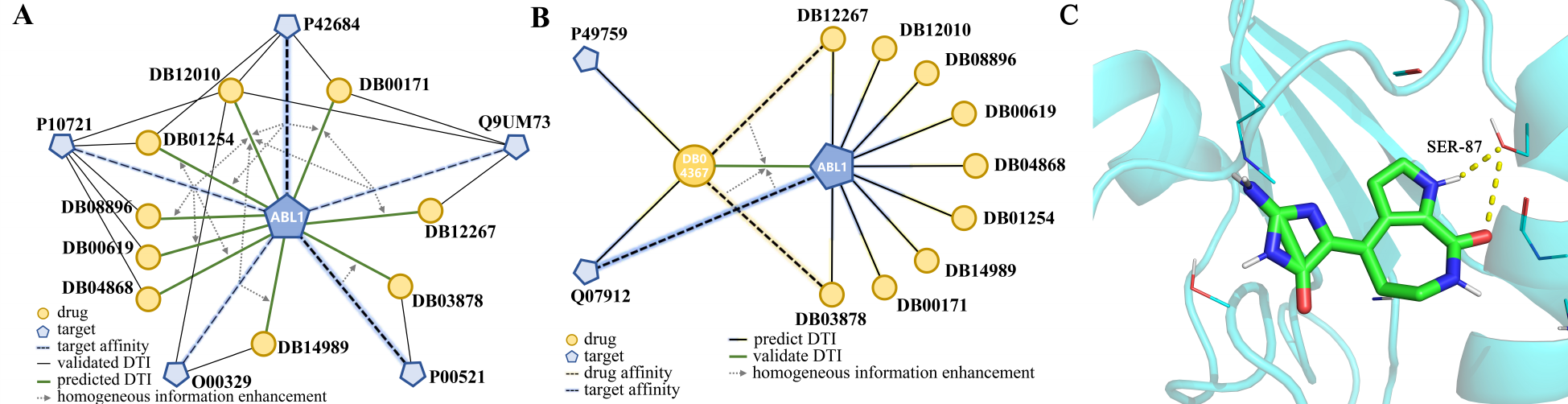}
\caption{Case study analysis. (A) SOC-DGL predicts the top 9 drugs targeting ABL1 and demonstrates the associations among them. (B) Verification of ADGL-Edge and EDGL-Edge between DB04367 and ABL1. (C) Perform a molecular docking experiment between DB04367 and ABL1.}
\label{fig:6}
\end{figure*}


\subsection{\textcolor{red}{Case study for the ABL1 in Chronic Myeloid Leukemi}} 
To assess the potential of SOC-DGL in discovering DTI in real-world applications, this study conducted a case study on Chronic Myeloid Leukemia (CML), a myeloproliferative neoplasm accounting for around 15\% of newly diagnosed adult leukemia cases~\cite{amin2017american}. The primary pathogenesis of CML is driven by the fusion of the ABL1 gene on chromosome 9, a tyrosine kinase, with the Breakpoint Cluster Region (BCR) gene on chromosome 22, resulting in the formation of a malignant target that initiates and progresses the disease~\cite{jabbour2018chronic}. The onset of CML is closely associated with ABL1 mutations, which lead to excessive cell proliferation, making it a key pathogenic factor in the disease~\cite{k2021protein}.

In this case study, the edges associated with drugs and ABL1 were removed to predict drugs targeting tyrosine protein kinase ABL1. The predicted DTI were ranked based on interaction scores, and the top 10 drugs were selected. Among these, 9 drugs were experimentally confirmed to target ABL1, as detailed in Table~\ref{tab:3}, where ADGL-Edge represented the direct similarity edges learned through ADGL, and EDGL-Edge represented the higher-order similarity edges constructed by EDGL. Extensive friend and higher-order friend relationships, learned through ADGL and EDGL, were observed between these drugs, as shown in Fig.~\ref{fig:6} (A). This demonstrated that ADGL and EDGL maintained strong biological interpretability when applied to real-world tasks. Two examples were provided to illustrate this. (1) For the third-ranked drug DB03878, ADGL learned high similarity information between ABL1 and P00521 via extensive friendship and high heterogeneity between P00521 and DB03878, predicting a potential association between DB03878 and ABL1. (2) For drug DB12010, SOC-DGL first learned four direct similarity relationships between ABL1 and P10721, P42684, Q9UM73, and O00329 through the ADGL module. Then, using EDGL, higher-order friendships were established, with DB12010 acting as an intermediary to create six (C$_{4}^{2}$) indirect friend channels, indirectly enhancing the higher-order similarity information between P10721, P42684, Q9UM73, and O00329, thus learning a potential association between ABL1 and DB01254.

In particular, the 10th drug DB04367 is a cyclin-dependent kinase inhibitor with pharmacological activity in the treatment of osteoarthritis and Alzheimer's disease~\cite{song2011cellular}.  However, no database or publication confirmed the direct association between DB04367 and ABL1. In this study, SOC-DGL suggested a possible link between them. As shown in Fig.~\ref{fig:6} (B) and Table.~\ref{tab:3}, further investigation revealed that SOC-DGL first learned three direct similarity relations between DB04367, DB12267, and DB03878 through the ADGL module, as well as an interaction between DB04367 and Q07912. Subsequently, through the EDGL module, a friendship pathway was constructed using ABL1 as an intermediary, indirectly improving the higher-order similarity information between DB12267 and DB03878, and thus uncovering a potential connection between ABL1 and DB04367. 

To verify the affinity of DB04367 and ABL1 at the molecular level, we conducted molecular coupling experiments on them with Pymol and AutoDockTools, shown in Fig.~\ref{fig:6} (C). The results indicated that DB04367 formed intermolecular forces with an intermolecular hydrogen bond in ABL1, confirming the interaction between DB04367 and ABL1.

\begin{table}[t]
\tiny
\centering
\begin{threeparttable}
\caption{SOC-DGL Prediction of the Top 10 Drugs Targeting ABL1.}
\begin{tabular}{
>{\centering\arraybackslash}p{0.05cm}
>{\centering\arraybackslash}p{1.37cm} 
>{\raggedright\arraybackslash}p{1.5cm} 
>{\raggedright\arraybackslash}p{1.5cm}
>{\centering\arraybackslash}p{0.95cm}
>{\centering\arraybackslash}p{0.95cm}}
\toprule
\textbf{No.} & \textbf{UniProtKB Entry} & \textbf{Name} & \textbf{Evidence} & \textbf{ADGL-Edge} & \textbf{EDGL-Edge}\\
\midrule
1 & DB12010 & Fostamatinib & Rolf et al.~\cite{rolf2015vitro} & 4 & 6\\
2 & DB00171 & ATP & Inming et al.~\cite{imming2006drugs} & 2 & 1\\
3 & DB03878 & NP-3-PP\tnote{(a)} & Berman et al.~\cite{berman2000protein} & 1 & 0\\
4 & DB08896 & Regorafenib & DrugBank & 1 & 0\\
5 & DB12267 & Brigatinib & Markham~\cite{markham2017brigatinib} & 1 & 0 \\
6 & DB14989 & Umbralisib & DrugBank & 1 & 0\\
7 & DB00619 & Imatinib & Haberler et al.~\cite{haberler2006immunohistochemical} & 1 & 0\\
8 & DB01254 & Dasatinib & Piccaluga et al.~\cite{piccaluga2007tyrosine} & 2 & 1\\
9 & DB04868 & Nilotinib & DrugBank & 1 & 0\\
10 & DB04367 & Debromohymenialdisine & None & 3 & 1\\
\bottomrule
\end{tabular}
\begin{tablenotes}
\item[(a)] Full name: N-[4-Methyl-3-[4-(3-Pyridinyl)-2-Pyrimidinyl]Amino]Phenyl]-3-Pyridinecarboxamide.
\end{tablenotes}
\label{tab:3}
\end{threeparttable}
\end{table}

\begin{table}[t]
\centering
\caption{Perform Similarity Ensemble Approach on the drug DB04367.}
\tiny
\begin{tabular}{
>{\centering\arraybackslash}p{1.2cm}
>{\centering\arraybackslash}p{1cm}
>{\raggedright\arraybackslash}p{3cm}
>{\centering\arraybackslash}p{0.8cm}
>{\centering\arraybackslash}p{0.75cm}}
\toprule
\textbf{Target Key} & \textbf{Target Name} & \textbf{Description} & \textbf{P-Value} $\downarrow$ & \textbf{MaxTc} $\uparrow$ \\
\midrule
CHK2\_HUMAN & CHEK2 & Serine/threonine-protein kinase Chk2 & 1.422e-09 & 0.62\\
CHK1\_HUMAN & CHEK1 & Serine/threonine-protein kinase Chk1 & 1.826e-09 & 0.62\\
UBP7\_HUMAN & USP7 & Ubiquitin carboxyl-terminal hydrolase 7 & 1.315e-07 & 0.63\\
\bottomrule
\label{tab:4}
\end{tabular}
\end{table}

\textcolor{red}{To further verify the potential association between DB04367 and ABL1 from the perspective of biological pathways, we used the SEA (Similarity Ensemble Approach)~\cite{keiser2007relating} to predict the potential targets of DB04367. SEA is a method based on chemical structure similarity, which infers possible targets by comparing the chemical structure of the query compound with known compounds in curated databases. Results shown in Table~\ref{tab:4} suggest that DB04367 may interact with CHEK2, CHEK1, and USP7. These P-values indicate strong associations between the compound and known ligands of the predicted targets. Related papers indicate that the above predicted targets may be biologically associated with ABL1: (1) CHEK2 and ABL1 are potentially linked in the DNA damage response pathway, particularly through coordinated regulation in EWS-mediated alternative splicing~\cite{paronetto2011ewing}; (2) CHEK1 and ABL1 are co-enriched in multiple Gene Ontology biological processes related to RAD51, such as homologous recombination repair of DNA double-strand breaks and regulation of cell proliferation~\cite{redati2022expression}; and (3) USP7 has been shown to enhance the stability of the BCR-ABL1 fusion protein, thereby promoting the proliferation and drug resistance of CML cells~\cite{jiang2021suppression}. These findings provide biological evidence for the potential association between DB04367 and ABL1, supporting the reliability of SOC-DGL's prediction.}


\section{Conclusion}
~\textcolor{red}{This study proposes SOC-DGL, a novel graph learning framework for DTI prediction. Inspired by real-world social interactions, SOC-DGL introduces ADGL to capture direct global associations and similarities between drugs and targets through extensive social connections. Additionally, EDGL employs a balance theory-driven even-polynomial graph filter to extract higher-order similarities from the heterogeneous drug-target graph, reflecting complex, higher-order interactions. Experimental results on four benchmark datasets demonstrate that SOC-DGL outperforms SOTA methods under both balanced and imbalanced conditions. Notably, SOC-DGL's successful prediction of the DTI between DB01254 and ABL1 emphasizes the importance of homogeneous information aggregation in uncovering complex indirect relationships in biological networks. We anticipate that future work will incorporate additional biological knowledge or advanced data augmentation techniques to address dataset noise and bias.
}


\title{Supplementary Materials}
\maketitle

\section{Appendix}

\subsection{Datasets}
To comprehensively evaluate the performance and applicability of the proposed method, we select four representative benchmark datasets: the KIBA dataset, the Davis dataset, the BindingDB dataset, and the DrugBank dataset.

\subsubsection{KIBA dataset}
In the KIBA dataset, some drugs and proteins do not have direct affinity metrics. However, it still contains comprehensive bioactivity measurement data between 2,111 kinase inhibitors and 229 kinases, including but not limited to key parameters such as K(d), K(i), and IC50. Notably, DTI are considered to have binding affinity when the K(d) value is below 30 units. To further optimize and utilize this dataset, we employ the Therapeutics Data Commons (TDC), a drug discovery application tool developed by Huang et al.~\cite{huang2022artificial} in 2022. Through the download and filtering capabilities of the TDC tool, we successfully obtain a refined dataset covering 1,720 drugs and 220 targets. This binary processed DTI dataset comprises 22,154 DTI, providing rich information and a solid foundation for our research and model training.

\subsubsection{Davis dataset}
The Davis dataset is a highly valuable resource in the field of DTI research, meticulously documenting the K(d) values obtained from wet laboratory tests between 72 kinase inhibitors and 442 kinases. This dataset covers over 80\% of the human catalytic protein kinase family, providing a wealth of information for understanding the bioactivity of kinase inhibitors. To further enhance the practicality of the data and the accuracy of analysis, we employ the Therapeutics Data Commons tool to process the Davis dataset. Through TDC’s advanced filtering and analysis capabilities, we successfully refine 7,320 positive samples, which encompass 68 drugs and 379 targets. This binary processed DTI dataset offers a more precise and comprehensive collection of positive samples for bioinformatics research, improving the quality of the data.

\subsubsection{BindingDB dataset}
BindingDB is a public dataset specifically designed to store experimentally measured binding affinities between small molecules (ligands) and their corresponding target biomolecules (targets). With an extensive dataset encompassing K(d) values for 10,665 drugs and 1,413 targets, BindingDB provides researchers with a valuable platform for gaining insights into DTI. To further optimize this data for specific research needs and enhance its usability, we employ the TDC tool and retain 9,166 DTI between 3,400 drugs and 886 targets.

\subsubsection{DrugBank dataset}
In ensuring the accuracy and reliability of the dataset, we implement a series of stringent quality control measures. Specifically, we conduct a thorough review of the drug molecules in the dataset, manually filtering out those drug sequences that cannot be effectively recognized through the Simplified Molecular Input Line-Entry System (SMILES) format using RDKit~\cite{landrum2019rdkit}. To address the sparsity issue of the DTI dataset, we further optimize the dataset by retaining only those drugs that interact with at least two targets. After these carefully designed selection processes, we ultimately obtain a dataset comprising 6,871 verified DTI, covering 1,821 drugs and 1,447 targets.

\subsection{Methodological Supplement}
\subsubsection{Detailed Process of Multi-view Learning}
Taking the application of multi-view learning in drug affinity matrices as an example, the specific process of multi-view learning can be detailed as follows. First, we respectively represent the MACC feature matrix, the Morgan feature matrix and the TOP feature matrix as $Y_1 \in \mathbb{R}^{v_1 \times n}$, $Y_2 \in \mathbb{R}^{v_2 \times n}$, and $Y_3 \in \mathbb{R}^{v_3 \times n}$, where $v_1$ represents the dimensionality of the MACC drug features, $v_2$ represents the dimensionality of the Morgan drug features, $v_3$ represents the dimensionality of the TOP drug features, and $n$ represents the number of drugs. Consequently, the combined drug feature matrix $Y$ can be obtained, expressed as follows:

\begin{equation}
Y = 
\begin{bmatrix}
Y_1^T \\
Y_2^T \\
Y_3^T
\end{bmatrix}
\in \mathbb{R}^{(v_1 + v_2 + v_3) \times n}
\end{equation}

Then, we define $v = v_1 + v_2 + v_3$ to represent the total number of perspectives. Following this, we introduce four auxiliary matrices $A$, $C1$, $C2$, and $C3$, each of which is a three-dimensional matrix of size $v \times n_d \times n_d$. These four auxiliary matrices encompass $v$ perspectives (distinct feature sets within multi-view learning), with each perspective having matrices of identical row and column dimensions. Specifically, $A$ denotes the relationship matrix among the various views, $C1$ is the auxiliary matrix for different views after singular value thresholding, $C2$ is the auxiliary matrix for different views following sparsity processing, and $C3$ is the auxiliary matrix that integrates information from multiple views. These auxiliary matrices are initialized to zero. Concurrently, we incorporate four Lagrange multiplier matrices \textcolor{red}{$\lambda1 \in \mathbb{R}^{(v+1)\times n_d \times n_d}$, $\lambda2 \in \mathbb{R}^{v \times n_d \times n_d}$, $\lambda3 \in \mathbb{R}^{v \times n_d \times n_d}$, and $\lambda4 \in \mathbb{R}^{v \times n_d \times n_d}$}, and set five hyperparameters $\mu$, $\beta_1$, $\beta_2$, $\lambda$, and $\mu_{\text{max}}$, with only the parameter $\mu$ being continuously updated. For all formulas in this section, the subscript $i$ signifies the index of a particular view, where $i = 0, 1, \ldots, v$. We define $C_{\text{sum}}$, which represents the sum of the values of the $C2$ matrices of all views except view $i$ itself.

\begin{equation}
C_{\text{sum}_i} = \sum_{i' \ne i} \text{C2}_{i'}
\end{equation}

Then, we update the auxiliary matrix ${A}$ iteratively. It updates the main relationship matrix ${A}$ of each view by integrating the auxiliary matrices of all views and the Lagrange multiplier matrices. The formula is as follows:
\begin{align}
{A}_i =\; & \mu_i (K + 3I)^{-1} \cdot \Big( \mu_i (K + C1 + C2 + C3) \notag \\
& - (\lambda2 + \lambda3 + \lambda4) + Y_i^T \lambda1 \Big)
\label{eq:A_update}
\end{align}
where $K = Y_i^T Y_i$, and $I$ is the identity matrix. Subsequently, we update the auxiliary matrix $C1$ iteratively and perform singular value thresholding on it. This process can effectively reduce the dimensionality and denoise the matrix. We perform singular value decomposition:
\begin{equation}
{A}_i + \frac{\lambda3}{\mu_i} = U_i \Sigma_i V_i^T
\end{equation}
where $U_i$ and $V_i$ are orthogonal matrices, and $\Sigma_i$ is the diagonal matrix containing the singular values. Then, the singular value matrix $\Sigma_i$ is thresholded to obtain $\Sigma_i' = \max(\Sigma_i - \frac{\beta_1}{\mu_i}, 0)$. Using the thresholded singular value matrix to reconstruct the matrix $C1$, we get $C1_i = U_i \Sigma_i' V_i^T$, and this completes the iterative update for $C1$. Next, we update $C2$ through soft thresholding, with the formula as follows:
\begin{equation}
C2 = \max({A} + \frac{\lambda2}{\mu_i} - \frac{\beta_2}{\mu_i}, 0) + \min({A} + \frac{\lambda2}{\mu_i} + \frac{\beta_2}{\mu_i}, 0)
\end{equation}

This operation updates the auxiliary matrix $C2$ for each view, effectively making the matrix sparser. Then, by integrating information from other matrices, we update the current matrix $C3$ to promote consistency and convergence, with the formula as follows:
\begin{equation}
C3_i = \frac{2 \lambda (v - 1) C_{\text{sum}_i} + \mu_i {A}_i + \lambda4_i}{2 \lambda (v - 1) + \mu_i}
\end{equation}
Subsequently, we adjust the Lagrange multipliers incrementally through the differences in constraint conditions, ensuring that the final solution satisfies all constraint conditions. The formula for this is as follows:
\begin{equation}
\left\{
\begin{aligned}
\lambda1 &= \lambda1 + \mu_i (Y_i - X_i {A}_i) \\
\lambda2 &= \lambda2 + \mu_i ({A}_i - C2_i) \\
\lambda3 &= \lambda3 + \mu_i ({A}_i - C1_i) \\
\lambda4 &= \lambda4 + \mu_i ({A}_i - C3_i)
\end{aligned}
\right.
\end{equation}

In the iterative optimization algorithm, convergence checking is an important step to determine whether the algorithm has achieved the expected precision, thus allowing the iteration to be terminated. Specifically, the goal of convergence checking is to ensure that the differences (errors) between various matrices are within an allowable error threshold. In the model, convergence checking is implemented by calculating the error values between different matrices, with the formulas as follows:
\begin{equation}
\left\{
\begin{aligned}
err_1 &= \max \left| {A}_i - C1_i \right| \\
err_2 &= \max \left| {A}_i - C2_i \right| \\
err_3 &= \max \left| {A}_i - C3_i \right| \\
err_4 &= \max \left| {A}_{i-1} - {A}_i \right|
\end{aligned}
\right.
\end{equation}

If all errors are less than the threshold value $\varepsilon = 0.000001$, the algorithm is considered to have converged. It should be noted that during one iteration, $i$ starts from zero and gradually increases to the number of views $v$, with matrix updates performed at each step. That is, in one iteration, the matrices mentioned earlier are all updated $v$ times. When one iteration ends, $i$ is reset to zero and the process of incrementing and updating matrices begins again. The iteration stops after 100 iterations or if the algorithm does not converge. As previously mentioned, $\mu$ is also updated with each iteration. In the iterative optimization algorithm, updating $\mu$ serves to control the weight of different terms in the objective function, allowing for a gradual approach to the optimal solution with each iteration. Specifically, $\mu$ is a regularization parameter that is updated according to certain rules in each iteration. We set a maximum value $\mu_{\text{max}}$, and an increment factor $\rho$ to update the value of $\mu$:
\begin{equation}
\mu_{t+1} = \min(\rho \mu_t, \mu_{\text{max}})
\end{equation}
where $\rho$ is a constant greater than 1, and $\mu_{\text{max}}$ is the upper limit of $\mu$. Once all iterations are completed, we obtain the final auxiliary matrix $C2$ after the iterative updates. We then define a matrix $C_{\text{avg}}$, which represents the average of the $C2$ matrices from all views:
\begin{equation}
C_{\text{avg}} = \frac{1}{v} \sum_j C2_j
\end{equation}

Ultimately, through pairwise multi-view learning on multiple drug features, we derive the drug affinity matrix $A_{DD} = |C_{\text{avg}}| + |C^{T}_{\text{avg}}|$. Similarly, we can obtain $A_{TT}$ using the aforementioned method. It is important to note that we use the iLearn software package to calculate features for the target sequences in these four different aspects (ACC, MORAN, CDT, PAAC), hence the number of views at this time is $v = v_1 + v_2 + v_3 + v_4$. Subsequently, we apply the min-max normalization method to both $A_{DD}$ and $A_{TT}$.
\subsubsection{Residual Connection Techniques}
\textcolor{red}{The corresponding formula is as follows:
\begin{equation}
\hat{H} = \omega H' + (1 - \omega) H''
\label{eq:H_prime}
\end{equation}
where \( \omega \) represents the weight, which is an adjustable hyperparameter. The integrated features \( \hat{H} \) will serve as the input for the decoder, used to predict DTI. Through residual connection feature fusion, this study gains a multifaceted understanding of the complex relationships between drugs and targets.}

\subsection{\textcolor{red}{The details of Baseline Methods}}
To validate the effectiveness and superiority of SOC-DGL, we conducted comparative experiments with seven leading methods that were widely recognized in DTI prediction. These models included GraphDTA, LRSpNM, MLMC, MULGA, MSI-DTI, BCMMDA and MMDG-DTI. 

\subsubsection{GraphDTA} GraphDTA, proposed by Thin Nguyen et al. in 2021, is a graph neural network-based model for predicting drug-target binding affinity. It represents drug molecules as graphs, using graph convolution to capture their spatial and chemical properties, improving DTI prediction accuracy by integrating various drug and target features.

\subsubsection{LRSpNM} LRSpNM, introduced by Gaoyan Wu et al. in 2021, predicts DTI using similarity information and dynamic pre-filling of unknown interactions. It employs low-rank matrix approximation, loss functions, and Laplacian regularization, utilizing the ADMM algorithm for efficient computation on large bioinformatics datasets.

\subsubsection{MLMC} MLMC, proposed by Yixin Yan et al. in 2022, predicts drug-disease associations by integrating multi-view learning for similarity information and matrix completion to fill missing values in the association matrix, improving the model’s predictive power for unknown drug-disease pairs.

\subsubsection{MULGA} MULGA, proposed by Jiani Ma et al. in 2023, is a multi-view graph autoencoder framework for DTI and drug repurposing prediction. It learns affinity matrices and infers missing interactions using a novel “guilt-by-association” negative sampling method, improving predictive accuracy.

\subsubsection{\textcolor{red}{MSI-DTI}} 
\textcolor{red}{MSI-DTI, proposed by Zhao et al. in 2024, utilizes multi-source information for DTI prediction. By integrating multi-head self-attention within a transformer architecture, it effectively captures complex interactions across diverse drug and target data sources, improving prediction accuracy and robustness.}

\subsubsection{\textcolor{red}{BCMMDA}} \textcolor{red}{BCMMDA, introduced by Huang et al. in 2024, enhances drug–protein interaction prediction by leveraging branch-chain mining (BCM) for fragment extraction and a multi-dimensional attention mechanism. This method improves feature representation and prediction accuracy through deep convolutional neural networks (CNNs) and attention-based refinement.}

\subsubsection{MMDG-DTI} MMDG-DTI, proposed by Yang Hua et al. in 2025, addresses domain-specific biases in existing models by using pre-trained LLMs to extract generalized features and a hybrid GNN for structural learning. It enhances generalization through Domain Adversarial Training, contrastive learning, and multi-modal feature fusion.

\subsection{\textcolor{red}{Metrics}}
\textcolor{red}{To comprehensively assess SOC-DGL's capability in handling drug-target repurposing, we set up five metrics: ACC, Precision, Recall, Specificity, and F1\_score.
\begin{equation}
\text{ACC} = \frac{TP + TN}{TP + TN + FN + FP}
\end{equation}
\begin{equation}
\text{Precision} = \frac{TP}{TP + FP}
\end{equation}
\begin{equation}
\text{Recall} = \frac{TP}{TP + FN}
\end{equation}
\begin{equation}
\text{Specificity} = \frac{TN}{TN + FP}
\end{equation}
\begin{equation}
\text{F1\_score} = \frac{2 \cdot \text{Precision} \cdot \text{Recall}}{\text{Precision} + \text{Recall}}
\end{equation}
where TP, FP, FN, and TN represent the number of samples that are actually positive and predicted as positive (True Positive Samples), actually negative and predicted as positive (False Positive Samples), actually positive and predicted as negative (False Negative Samples), and actually negative and predicted as negative (True Negative Samples), respectively. Here, "positive" indicates that there is an association between a drug and a target, while "negative" indicates that there is no association (or very low association) between a drug and a target. Subsequently, based on the aforementioned metrics, we plot the ROC curve and the PR curve, and then calculate the area under the ROC curve (AUROC) and the area under the PR curve (AUPR). The five metrics mentioned earlier, along with the two curve area metrics, provide a total of seven indicators that can measure the model’s performance from different aspects, assisting in the selection and optimization of classification models to meet the needs of DTI prediction.}

\textcolor{red}{In drug discovery and bioinformatics, selecting negative samples is crucial. Positive samples are drugs known to interact with targets, while negative samples are those with unknown interactions. To improve model performance, we employ an association-based negative sampling method that selects more reliable negative samples by leveraging drug similarity. If a drug’s neighbors (high-similarity drugs) interact with a target, it is likely to do so as well. Conversely, drugs with low similarity to known positives are less likely to share common targets and can be considered negative samples, reducing the model’s bias toward positive interactions.}

\subsection{Validation experiment on imbalanced loss function in imbalanced datasets}

The formulas for the four loss functions (SLF, FLF, WLF, and RLF) are as follows:
\begin{equation}
\mathcal{L}_S = -\frac{1}{n_d\cdot n_t}(\!\!\sum_{(i,j)\in y^+} \!\!\!\!\log(h_{ij}^*)+\!\!\!\!\sum_{(i,j)\in y^-}\!\!\!\!\log(1-h_{ij}^*)) 
\end{equation}
\begin{equation}
\mathcal{L}_F = - \frac{|y^-|}{|y^+|} \left( 1 - h_{ij}^* \right)^\gamma \log(h_{ij}^*)
\end{equation}
\begin{equation}
\mathcal{L}_W = -\frac{1}{n_d\cdot n_t}(\frac{|y^-|}{|y^+|}\!\!\sum_{(i,j)\in y^+} \!\!\!\!\log(h_{ij}^*)+\!\!\!\!\sum_{(i,j)\in y^-}\!\!\!\!\log(1-h_{ij}^*))
\end{equation}
\begin{equation}
\mathcal{L}_R = -\frac{1}{n_d\cdot n_t}(\varpi\frac{|y^-|}{|y^+|}\!\!\sum_{(i,j)\in y^+} \!\!\!\!\log(h_{ij}^*)+\!\!\!\!\sum_{(i,j)\in y^-}\!\!\!\!\log(1-h_{ij}^*))
\end{equation}
where \( \mathcal{L}_S \), \( \mathcal{L}_F \), \( \mathcal{L}_W \), and \( \mathcal{L}_R \) represent the SLF, FLF, WLF, and RLF, respectively. \( n_d \) denotes the number of drugs in the dataset, \( n_t \) is the number of targets, \( |y^+| \) is the count of positive samples, and \( y^-| \) is the count of negative samples. \( h_{ij}^* \) represents the predicted interaction probability between drug \( i \) and target \( j \). The adjustment parameter \( \gamma \) modulates the weight of easily classified samples, and \( \varpi \) adjusts the weight of negative samples without affecting positive sample weights.

\section*{References}

\vspace{-1.5em}
\bibliographystyle{IEEEtran}
\bibliography{reference}

\end{document}